\documentclass[10pt, a4paper]{article}
\usepackage{booktabs}
\usepackage{array}
\usepackage[final]{lrec2026} 
\newfontfamily\greekfont{FreeSans.ttf}
\usepackage{tikz}
\usetikzlibrary{shapes.geometric, arrows, positioning, shadows}
\usepackage{subcaption}
\usepackage{graphicx}
\usepackage{float} 
\usepackage{pgfplots} 
\pgfplotsset{compat=1.18}
\usepackage{pgf-pie} 

\title{Ancient Greek to Modern Greek Machine Translation: A Novel Benchmark and Fine-Tuning Experiments on LLMs and NMT Models}

\name{Spyridon Mavromatis$^{\ast \dagger}$, Sokratis Sofianopoulos$^{\dagger}$, Prokopis Prokopidis$^{\dagger}$, Maria Giagkou$^{\dagger}$}

\address{$^{\ast}$National and Kapodistrian University of Athens, Department of Informatics and Telecommunications \\
         $^{\dagger}$Institute for Language and Speech Processing, Athena RC \\
         spyros.mauromatis@athenarc.gr, s\_sofian@athenarc.gr, prokopis@athenarc.gr, mgiagkou@athenarc.gr\\}

\abstract{
Machine Translation (MT) for Ancient Greek (AG) to Modern Greek (MG) is a low-resource task, constrained by the lack of large-scale, high-quality parallel data. We address this gap by introducing the AG-MG Parallel Corpus, a new resource containing 132,481 sentence-aligned pairs derived from literary, historical, and biblical texts. We present a novel corpus creation pipeline that combines web-scraped, excerpt-level data with a multi-stage sentence-level alignment, and refinement process. Our method uses VecAlign with LaBSE embeddings, which we first fine-tune on a manually-aligned AG-MG subset, followed by an LLM-based error/misalignment correction phase using Gemini 2.5 Flash to ensure high alignment quality. Furthermore, we provide the first comprehensive benchmark of modern MT models on this task, evaluating three fine-tuning strategies across NMT models (NLLB, M2M100) and a Greek LLM (Llama-Krikri-8B). Our experiments show that fine-tuning yields significant improvements over base models, increasing performance by up to +10.3 BLEU points. Specifically, full-parameter fine-tuning of Llama-Krikri-8B achieves the highest overall performance with a BLEU score of 13.16, while the QLoRA-adapted M2M100-1.2B model demonstrates the largest relative gains and highly competitive results. Our dataset and models represent a significant contribution to Greek NLP.
 \\ \newline  }

\begin{document}

\maketitleabstract

\section{Introduction}
\label{sec:introduction}

Machine Translation (MT) has evolved from rule-based and statistical approaches to neural sequence-to-sequence models and is now being reshaped by Large Language Models (LLMs). Recent studies suggest that LLMs, pre-trained on extensive multilingual data, can rival or complement traditional encoder–decoder MT, particularly in settings with limited resources where parallel data is scarce \citep{lyu2023paradigm}. This trend has led to revisiting under-resourced, stylistically rich language pairs such as Ancient Greek (AG) to Modern Greek (MG), where the source side spans multiple dialects, historical periods, and genres. 

A critical bottleneck in machine translation of AG$\rightarrow$MG is the lack of large, sentence-aligned parallel corpora. Historical resources often exist at excerpt or document level and contain a lot of noise, such as editorial notes, markup, and orthographic variants. Recent work shows that neural embedding–based aligners can substantially improve sentence alignment accuracy and scalability on classical languages. For example, VecAlign\footnote{\url{https://github.com/thompsonb/vecalign}} combined with multilingual sentence embeddings (e.g., LaBSE\footnote{\url{https://huggingface.co/sentence-transformers/LaBSE}} or LASER\footnote{\url{https://github.com/facebookresearch/LASER}}) has been found effective on Ancient Greek and Latin texts, outperforming traditional approaches (e.g., Hunalign) and handling very long documents \citep{craig2023testing,thompson2019vecalign,feng2020language,yousef2022automatic}. These developments pave the way for the creation of higher-quality AG$\rightarrow$MG datasets suitable for MT.

At the modeling level, two families of systems are promising for Greek: (i) multilingual encoder–decoder NMT models such as M2M100 \citep{fan2021beyond} and NLLB \citep{costa2022no}, and (ii) open Greek LLMs, notably Llama–Krikri\footnote{\url{https://huggingface.co/ilsp/Llama-Krikri-8B-Instruct}} , which offer strong Greek fluency and tokenization while remaining accessible for adaptation \citep{roussis2025krikri,voukoutis2407meltemi}. This paper presents a high-quality AG$\rightarrow$MG sentence-level parallel corpus, as well as fine-tuning NMT models and a Greek LLM for this task.

In this work, we address the resource gap and we provide a systematic evaluation of modern translation models. Our key contributions include the following:

\begin{enumerate}
    \item We introduce the \textbf{AG-MG Parallel Corpus}, the largest sentence-aligned corpus for this low-resource language pair, containing 132,481 high-quality aligned sentence pairs. The corpus is enriched with extensive metadata, including author, title, segment index, translator, ancient Greek dialect, genre, and era. An excerpt/paragraph-level version of the corpus will also be created soon.
    
    \item We present a \textbf{novel hybrid alignment pipeline} for creating the corpus. Our method first uses a domain-adapted LaBSE model for initial alignment and then leverages a LLM (Gemini 2.5 Flash developed by \citealp{comanici2025gemini}) for misalignment detection and correction, ensuring superior alignment quality.
    
    \item We conduct the \textbf{first large-scale benchmark} of both NMT and LLM-based models for AG-to-MG translation. We compare three different fine-tuning strategies (Full-parameter, LoRA, and QLoRA) on NLLB\footnote{\url{https://huggingface.co/docs/transformers/en/model_doc/nllb}}, M2M100\footnote{\url{https://huggingface.co/docs/transformers/en/model_doc/m2m_100}} and Llama-Krikri-8B, providing a clear picture of the current state-of-the-art on this task.
\end{enumerate}

Our results demonstrate that fine-tuned models significantly outperform their base versions (yielding improvements of up to +10.3 BLEU points). The fully fine-tuned Llama-Krikri-8B LLM achieved the highest overall BLEU score (13.16), while the M2M100-1.2B model showed the largest relative gains. This highlights the potential of both specialized LLMs and adapted NMT architectures for this task. We hope that this study will facilitate future research in Greek NLP and Digital Humanities.

\section{Related Work}
\label{sec:related}

Our work is positioned at the intersection of classical language resource creation and low-resource machine translation.

\subsection{Parallel Corpora and Alignment}
Although there are monolingual corpora for Ancient Greek, such as the Diorisis corpus \citeplanguageresource{vatri2018diorisis}, the creation of large-scale parallel corpora for MT is a more recent challenge. Efforts have focused on establishing gold standards \citep{yousef-etal-2022-automatic,palladino2023translation} and evaluating alignment methods for classical languages \citep{yousef2022automatic,craig2023testing,yousef2023classical,sommerschield2023machine}.

A key finding from this line of work is that neural, embedding-based aligners significantly outperform older, traditional methods (e.g., length-based heuristics or tools like Hunalign). \citet{craig2023testing} conducted a detailed evaluation of various tools specifically for \textbf{sentence-level alignment} on AG and Latin texts. Their results demonstrated that a combination of VecAlign and LaBSE embeddings provides the most accurate sentence alignments, especially for long and literary documents common in classical studies.

\subsection{Machine Translation of Ancient Greek}
Machine Translation involving Ancient Greek remains a definitive low-resource task. The majority of published research has focused on translating AG to English (AG$\rightarrow$EN). \citet{kolovou2024machine}, for instance, experimented with OpenNMT on an AG$\rightarrow$EN dataset, highlighting the challenges of the task. More recently, \citet{rapacz2025low} improved neural models for interlinear translation by incorporating morphological information to better handle the rich inflection of Ancient Greek, focusing on translation into English and Polish (AG$\rightarrow$EN/PL).

To our knowledge, the AG$\rightarrow$MG language pair is significantly less explored and no large-scale benchmark exists that compares modern, high-capacity NMT models (NLLB, M2M100) and new Greek LLMs (Llama-Krikri) for this task \citep{sommerschield2023machine,papantoniou2024nlp}. This study seeks to address this gap by providing the first comprehensive evaluation of these models on a new, high-quality dataset.

\section{The Ancient-Modern Greek Parallel Corpus}
\label{sec:corpus}

We introduce the AG-MG Parallel Corpus, a new sentence-aligned dataset designed for AG$\rightarrow$MG machine translation. This section details the data sources, our alignment methodology, and the final corpus characteristics.

\subsection{Data Sources}
The corpus aggregates parallel texts from three main types of digital resources, aiming for diversity in genre, dialect, and period of Ancient Greek:
\begin{itemize}
    \item \textbf{Digitized Anthologies:} Collections focusing primarily on classical Attic prose (approx. 5th--4th century BC), offering curated AG texts alongside scholarly MG translations.
    \item \textbf{Comprehensive Digital Libraries:} Larger repositories of Ancient Greek literature containing a wider variety of AG dialects (e.g., Homeric/Epic from the 8th century BC, Ionic, Doric, Attic and later Hellenistic Koine) paired with MG translations. While some translations in these sources date to the early 20th century, we strictly selected only contemporary, monotonic Modern Greek versions, explicitly excluding archaic or Katharevousa translations.
    \item \textbf{Biblical Texts Resources:} Digital versions of the Holy Bible providing AG texts (Septuagint/LXX and New Testament, representing Hellenistic Koine from the 3rd century BC to the 6th century AD) aligned at the verse level with contemporary MG translations.
\end{itemize}
These sources typically provide texts aligned at the excerpt or paragraph level, necessitating the sentence alignment pipeline described below.

\subsection{Preprocessing and Alignment Pipeline}
To create a high-quality, sentence-aligned corpus, we implemented a multi-stage pipeline:

\paragraph{1. Scraping and Initial Cleaning:} We first scraped the public web sources using standard Python libraries (BeautifulSoup\footnote{\url{https://beautiful-soup-4.readthedocs.io/en/latest/}} and Scrapy\footnote{\url{https://scrapy.org/}}), extracting AG texts, MG translations, and available metadata (author, title, translator, etc.) into JSONL format. Basic cleaning removed residual HTML tags and markup.

\paragraph{2. Deep Cleaning and Segmentation:} We applied a more thorough cleaning process to remove noise such as page numbers, editorial brackets, translator comments, and inconsistent punctuation. Both AG and MG texts were then segmented into sentences using the Stanza library\footnote{\url{https://stanfordnlp.github.io/stanza/}}.

\paragraph{3. Fine-Tuned Embedding Alignment:} For the non-Bible sources where sentence alignment was needed, we employed VecAlign, following prior work \citep{craig2023testing}. Crucially, instead of using off-the-shelf embeddings, we fine-tuned LaBSE \citep{feng2020language} on \textbf{1,000 manually aligned AG-MG sentence pairs}, varying in genre and ancient Greek dialect and extracted from our sources. This domain adaptation step significantly improved the embedding quality for the specific nuances of AG-MG alignment. VecAlign was then run using these fine-tuned embeddings to generate initial sentence alignments. For the Bible texts, alignment was straightforward using the existing verse indices.

\paragraph{4. LLM-Based Refinement:} While the fine-tuned VecAlign+LaBSE approach yielded good results, manual inspection revealed residual misalignments, particularly with non-literal translations or sentence splitting/merging. To ensure the highest possible quality of the corpus, we implemented a refinement step using the Gemini 2.5 Flash API \citep{comanici2025gemini}. We designed prompts\footnote{\textit{You are a cautious data editor. Your job is to fix alignment errors between Ancient Greek (“grc”) and Modern Greek (“ell”) sentences inside a JSONL file. Do not translate or paraphrase any text. Only merge/split and reorder within a row while keeping the original order of sentences.}} that instructed the model to evaluate the semantic equivalence of proposed AG-MG pairs from VecAlign and flag or correct likely misalignments (e.g., 1-to-many, many-to-1, many-to-many, or incorrect 1-to-1 mappings). This automated LLM-based verification proved highly effective in identifying and fixing errors. After applying the LLM refinement, we conducted a careful human evaluation on a subset of 150 aligned pairs and estimated that our dataset contains approximately 95\% correct alignments. The 5\% of alignment errors are mainly caused by inconsistent MG translations and translator comments.

\paragraph{5. Deduplication and Multi-Reference Handling:} Following \citet{lee2021deduplicating}, we performed deduplication on all splits based on the MG sentences to remove near-identical translation variants that might skew model training. However, drawing inspiration from multi-reference MT training \citep{zheng2018multi, khayrallah2020simulated}, when sources provided distinct translations from different translators for the same AG sentence, we retained these variants specifically within
the training set, enriching the corpus with translation and stylistic variation. 

\subsection{Data Card and Statistics}
The final corpus consists of \textbf{132{,}481} sentences per language (AG$\rightarrow$MG), with \textbf{2{,}311{,}448} AG tokens/words (avg.\ 17.45 per sentence) and \textbf{3{,}060{,}949} MG tokens/words (avg.\ 23.10 per sentence) as presented in Table~\ref{tab:stats}. Each dataset entry preserves metadata for \textit{author, title, segment index, url, translator, ancient Greek dialect, genre, and era} and enables comparisons to excerpt-level variants (kept separately for future excerpt/paragraph-level work).

\begin{table}[t]
\centering
\small
\setlength{\tabcolsep}{3.5pt}
\begin{tabular*}{\columnwidth}{@{\extracolsep{\fill}} l rr @{}}
\toprule
 & \textbf{Ancient Greek} & \textbf{Modern Greek} \\
\midrule
Sentences & 132{,}481 & 132{,}481 \\
Tokens / Words & 2{,}311{,}448 & 3{,}060{,}949 \\
Avg. tokens per sent. & 17.45 & 23.10 \\
\bottomrule
\end{tabular*}
\caption{Sentence and token statistics for the released sentence-level corpus.\protect\footnotemark}
\label{tab:stats}
\end{table}
\footnotetext{Tokens (or words) are counted as whitespace-separated items. The exact number of Ancient and Modern Greek sentences may be higher, since aligned sentence pairs can contain multiple sentences per side.}

\begin{table}[t]
\centering
\small
\setlength{\tabcolsep}{4pt} 
\renewcommand{\arraystretch}{1.1} 
\begin{tabular*}{\columnwidth}{@{\extracolsep{\fill}} >{\raggedright\arraybackslash}p{0.65\columnwidth} >{\centering\arraybackslash}m{0.25\columnwidth} @{}}
\toprule
\textbf{Split} & \textbf{Sentence Pairs} \\
\midrule
Train  & 128{,}231 \\
Dev    & 2{,}000 \\
Test (main reporting set) & 2{,}000 \\
Stress (rarer dialects: Ionic, Doric, Homeric) & 250 \\
\midrule
\textbf{Total} & \textbf{132{,}481} \\
\bottomrule
\end{tabular*}
\caption{Corpus splits for AG$\rightarrow$MG experiments.}
\label{tab:splits}
\end{table}

\begin{table}[t]
\centering
\small
\begin{tabular}{l r}
\toprule
\textbf{Ancient Greek Dialect} & \textbf{Sentences} \\
\midrule
Attic (Αττική) & 75{,}887 \\
Ionic (Ιωνική) & 8{,}725 \\
Doric (Δωρική) & 2{,}614 \\
Homeric / Epic (Ομηρική, Επική) & 6{,}729 \\
Hellenistic Koine (Ελληνιστική Κοινή) & 38{,}526 \\ 
\midrule
\textbf{Total} & \textbf{132{,}481} \\
\bottomrule
\end{tabular}
\caption{Distribution of aligned sentence pairs across major Ancient Greek dialects in the corpus.}
\label{tab:dialects}
\end{table}

\section{Experimental Setup}
\label{sec:experiments}

We evaluate the effectiveness of our AG-MG Parallel Corpus by fine-tuning several state-of-the-art NMT and LLM models. This section details the dataset splits, models, fine-tuning procedures, and evaluation metrics used.

\subsection{Dataset Splits}
We split the 132,481 sentence pairs as described in Table~\ref{tab:splits}. The training set comprises 128,231 pairs. We use separate development (Dev) and test sets, each containing 2,000 pairs, randomly sampled while ensuring no overlap with the training set or each other. Additionally, we created a small stress set of 250 pairs specifically containing rarer AG dialects (Ionic, Doric, Homeric) to evaluate model generalization beyond the dominant Attic and Koine dialects. A detailed breakdown of the exact dialectal composition for each of these splits is provided in Appendix \ref{app:statistics}.

\subsection{Models}
We benchmark two types of models:
\begin{itemize}
    \item \textbf{Multilingual NMT Models:} We selected strong open-source encoder-decoder models known for their ability to handle multiple languages: 
        \begin{itemize}
            \item facebook/nllb-200-distilled-600M \citep{costa2022no}
            \item facebook/nllb-200-distilled-1.3B \citep{costa2022no}
            \item facebook/m2m100\_1.2B \citep{fan2021beyond}
        \end{itemize}
    \item \textbf{Greek LLM:} We used ilsp/Llama-Krikri-8B-Instruct \citep{roussis2025krikri}, an open Greek LLM based on the Llama architecture, selected for its strong native Greek fluency.
\end{itemize}
For each model, we evaluate both its zero-shot (base) performance and its performance after fine-tuning.

\subsection{Fine-Tuning Strategies}
We employed three fine-tuning strategies to adapt the models to the AG$\rightarrow$MG task:
\begin{itemize}
    \item \textbf{Full Fine-Tuning (Full FT):} All model parameters were updated during training. Applied to M2M100-1.2B and Llama-Krikri-8B.
    \item \textbf{LoRA (Low-Rank Adaptation):} \citep{hu2022lora} A parameter-efficient fine-tuning (PEFT) method that injects trainable low-rank matrices into the model layers, freezing the original weights. Applied to NLLB-600M and NLLB-1.3B.
    \item \textbf{QLoRA (Quantized LoRA):} \citep{dettmers2023qlora} A more memory-efficient PEFT method combining 4-bit quantization with LoRA. Applied to M2M100-1.2B and Llama-Krikri-8B.
\end{itemize}

\subsection{Training Details}
Fine-tuning was performed using either Google Colab Pro (L4 GPU) or the CINECA\footnote{\url{https://docs.hpc.cineca.it/index.html}} supercomputing infrastructure (using 1 to 4 NVIDIA A100 64GB GPUs\footnote{The LEONARDO supercomputer features custom NVIDIA Ampere A100 accelerators equipped with 64GB of HBM2e memory, a specific hardware configuration designed for this EuroHPC system.}) for the larger models and full fine-tuning runs. Key hyperparameters are summarized in Table~\ref{tab:hparams}.

All experiments used the Hugging Face \texttt{transformers}\footnote{\url{https://huggingface.co/docs/transformers}} library, with \texttt{peft}\footnote{\url{https://huggingface.co/docs/peft}} for LoRA/QLoRA and \texttt{bitsandbytes}\footnote{\url{https://github.com/bitsandbytes-foundation/bitsandbytes}} for quantization. Training employed AdamW optimizer variants (\texttt{paged\_adamw\_8bit} or \texttt{adamw\_torch\_fused}) with a cosine learning rate scheduler and warmup. We used mixed precision (BF16 on A100s, FP16 on L4). For NLLB and M2M100 models, we expanded the tokenizer vocabulary with 148 (for the NLLB models) and 122 (for the M2M100 models)  Ancient Greek characters/tokens missing from the base models and resized the model embeddings accordingly, unfreezing them during PEFT to allow adaptation. The Full FT of Krikri-8B utilized \texttt{DeepSpeed ZeRO Stage 3} for distributed training across 4 A100s. Models were trained for 5 epochs, selecting the best checkpoint based on validation loss. 

\subsubsection{Vocabulary Adaptation and Smart Initialization}
A significant challenge in adapting multilingual NMT models like M2M100 and NLLB to Ancient Greek is related to how these models handle Ancient Greek tokens. Ancient Greek utilizes the Polytonic system, featuring diacritics such as \textit{oxia} (acute accent), \textit{varia} (grave accent), \textit{perispomeni} (circumflex accent), \textit{psili} (smooth breathing), \textit{dasia} (rough breathing) and more, which is largely absent from the pre-training data of modern multilingual models. Consequently, standard tokenizers treat many Polytonic characters as unknown tokens (\texttt{<unk>}), leading to input fragmentation or complete "input blindness", as observed in the base M2M100 model.

To overcome this, we implemented a robust four-stage vocabulary adaptation pipeline:

\begin{enumerate}
    \item \textbf{Token Discovery:} We scanned the entire Ancient Greek training corpus to identify characters that the base tokenizer could not resolve. This process revealed 122 missing Ancient Greek characters/tokens for the M2M100 and 148 for the NLLB models.
    
    \item \textbf{Dictionary Update:} These identified characters were explicitly added to the model's tokenizer, assigning them unique IDs and preventing them from being mapped to \texttt{<unk>}.
    
    \item \textbf{Embedding Resizing:} We structurally resized the model's input and output embedding matrices to accommodate the newly added token IDs.
    
    \item \textbf{Smart Initialization (Weight Transplant):} Initializing new embeddings with random noise can significantly slow down convergence, as the model must learn the semantic value of these characters from scratch. Instead, we employed a "smart initialization" strategy. For each new Polytonic character (e.g., \textit{{\greekfont ᾷ}}), we programmatically identified its closest base character in the existing vocabulary (e.g., \textit{{\greekfont α}}) by stripping diacritics. We then copied the pre-trained embedding weights of the base character to the new Polytonic token. This effectively provides the model with a head start, allowing it to treat the new character as a variation of a known vowel or consonant rather than a random symbol.
\end{enumerate}

This methodology ensured that fine-tuning focused on learning the syntactic and dialectal nuances of Ancient Greek rather than learning basic character representations, leading to the substantial performance gains observed in our experiments.

It is important to note that this vocabulary adaptation process was applied exclusively to the NMT models (M2M100 and NLLB). The LLM evaluated in this study (Llama-Krikri) utilizes tokenizer architectures with significantly larger vocabularies and byte-level fallback mechanisms (e.g., BPE or TikToken), which inherently handle rare Polytonic characters without requiring explicit vocabulary expansion.

\begin{table*}[t] 
\centering
\setlength{\tabcolsep}{4.5pt} 
\begin{tabular}{l l c c c c c c}
\toprule
\textbf{Model} & \textbf{Adaptation} & \textbf{Quant.} & \textbf{Rank (r)} & \textbf{Eff. Batch} & \textbf{LR} & \textbf{Platform} \\
\midrule
NLLB-600M & LoRA    & 8-bit & 16 & 16 & 2e-4 & L4 (Colab) \\
NLLB-1.3B & LoRA    & 8-bit & 16 & 12 & 1e-4 & L4 (Colab) \\
M2M100-1.2B & QLoRA   & 4-bit NF4 & 16 & 12 & 1e-4 & L4 (Colab) \\
M2M100-1.2B & Full FT & None  & N/A & 12 & 1e-4 & 1xA100 (CINECA) \\ 
Krikri-8B-IT & QLoRA   & 4-bit NF4 & 32 & 16 & 8e-5 & 1xA100 (CINECA) \\
Krikri-8B-IT & Full FT & None  & N/A & 64\footnotemark[18] & 2e-5 & 4xA100 (CINECA, DS-Z3) \\ 
\bottomrule
\end{tabular}
\caption{Key hyperparameters for fine-tuning experiments. Eff. Batch = per-device batch size $\times$ gradient accumulation steps $\times$ num GPUs. LR = Learning Rate. All experiments ran for 5 epochs.}
\label{tab:hparams}
\end{table*}
\footnotetext[18]{Calculated using DeepSpeed config: 2 (micro-batch) * 8 (grad accum) * 4 (GPUs).}

\subsection{Evaluation Metrics}
We evaluate translation quality using a comprehensive suite of standard metrics:
\begin{itemize}
    \item \textbf{BLEU:} \citep{papineni2002bleu} SacreBLEU implementation with \texttt{13a} and \texttt{intl} tokenization.
    \item \textbf{chrF/chrF++:} \citep{popovic2015chrf,popovic2017chrf++} Character n-gram metrics, with chrF++ using word n-grams (\texttt{word\_order=2}).
    \item \textbf{TER (Translation Edit Rate):} \citep{snover2006study} Measures the number of edits required to match the reference. Lower is better.
    \item \textbf{BERTScore:} \citep{zhang2019bertscore} Computes semantic similarity using contextual embeddings (using \texttt{xlm-roberta-large}). We report F1.
    \item \textbf{COMET:} \citep{rei2020comet} A neural metric trained to predict human judgments of translation quality (using \texttt{Unbabel/wmt22-comet-da}). Higher is better.
\end{itemize}
All metrics were calculated using standard implementations, comparing model outputs against the reference translations in the Test and Stress sets.

\section{Results and Analysis}
\label{sec:results}

We present the evaluation results on the Test and Stress sets, comparing the zero-shot (Base) performance of the pre-trained models against their fine-tuned versions across all metrics.

\subsection{Results on Test Set}
Table~\ref{tab:main_results_test} summarizes the performance of all models on the main Test set (2,000 pairs).

\begin{table*}[t] 
\centering
\setlength{\tabcolsep}{4pt} 
\begin{tabular}{ll rrr rrr r} 
\toprule
\textbf{Model} & \textbf{Method} & \textbf{BLEU$\uparrow$} & \textbf{chrF++$\uparrow$} & \textbf{TER$\downarrow$} & \textbf{BERTScore F1$\uparrow$} & \textbf{COMET$\uparrow$} & \textbf{$\Delta$BLEU} \\
\midrule
\multicolumn{8}{c}{\textit{NMT Models}} \\
\midrule
NLLB-600M & Base & 1.55 & 16.86 & 106.80 & 0.880 & 0.539 & - \\
          & + LoRA & 7.43 & 29.31 & 88.32 & 0.903 & 0.667 & +5.88 \\
\midrule
NLLB-1.3B & Base & 2.15 & 17.78 & 106.41 & 0.885 & 0.573 & - \\
          & + LoRA & 8.01 & 30.02 & 87.74 & 0.905 & 0.687 & +5.86 \\
\midrule
M2M100-1.2B & Base & 0.62 & 10.70 & 100.50 & 0.858 & 0.475 & - \\ 
            & + QLoRA & 10.96 & 33.09 & \textbf{82.99} & \textbf{0.911} & 0.710 & \textbf{+10.34} \\ 
            & + Full FT & 9.60 & 31.16 & 83.43 & 0.908 & 0.692 & +8.98 \\ 
\midrule
\multicolumn{8}{c}{\textit{LLM}} \\
\midrule
Krikri-8B-IT & Base & 8.29 & 29.87 & 88.13 & 0.895 & 0.695 & - \\ 
             & + QLoRA & 11.90 & 34.07 & 84.16 & 0.906 & \textbf{0.713} & +3.61 \\ 
             & + Full FT & \textbf{13.16} & \textbf{34.71} & 83.68 & 0.848 & 0.702 & +4.87 \\ 
\bottomrule
\end{tabular}
\caption{Main results on the Test set (2,000 sentence pairs). BLEU scores are BLEU-13a. $\Delta$BLEU shows improvement over the base model. Best score per metric in bold.}
\label{tab:main_results_test}
\end{table*}

The results clearly demonstrate the effectiveness of fine-tuning. All fine-tuned models achieve significant improvements over their base zero-shot counterparts across all metrics, with BLEU scores increasing by up to \textbf{+10.3} points for the QLoRA-adapted M2M100 model (using its corresponding base score as reference).

The Greek LLM, Llama-Krikri-8B-Instruct, emerges as the top overall performer. Its base performance (BLEU 8.29) is notably higher than the NMT base models, indicating strong inherent capabilities for Greek-related tasks. After full fine-tuning, it reaches the highest BLEU (13.16) and chrF++ (34.71) across all models. Furthermore, its QLoRA variant achieves the highest COMET score (0.713), suggesting high perceived translation quality.

 A striking observation concerns the performance of the base M2M100-1.2B, NLLB-600M and NLLB-1.3B models, which performed very poorly. The base M2M100-1.2B model recorded the lowest BLEU score (0.62) among all experiments. These near-zero performances are not due to poor grammar but rather a complete failure to process the Polytonic script. Since the base tokenizers treat most Ancient Greek characters (e.g., vowels with breathing marks and varia/oxia) as unknown tokens or noise, the models suffer from severe hallucinations. This validates the necessity of our vocabulary expansion and embedding resizing strategy, which unlocked the true potential of the models. While the base zero-shot performance of the M2M100-1.2B model is very low due to its limited exposure to Polytonic Greek, fine-tuning with QLoRA achieves a BLEU of 10.96, the lowest TER (82.99), and the highest BERTScore (0.911), showing a massive relative improvement (+10.34 $\Delta$BLEU). This observation highlights the paramount role of data quality in resource-constrained tasks. It suggests that a high-quality, curated dataset, such as our AG-MG corpus, contributes significantly to a fine-tuned model's translation efficacy, effectively mitigating the effects of a less capable base model.

\subsection{Results on Stress Set (Rare Dialects)}
Table~\ref{tab:stress_results} shows performance on the Stress set (250 pairs), designed to test generalization to rarer AG dialects (Ionic, Doric, Homeric).

\begin{table*}[t] 
\centering
\setlength{\tabcolsep}{4pt} 
\begin{tabular}{ll rrr rrr r} 
\toprule
\textbf{Model} & \textbf{Method} & \textbf{BLEU$\uparrow$} & \textbf{chrF++$\uparrow$} & \textbf{TER$\downarrow$} & \textbf{BERTScore F1$\uparrow$} & \textbf{COMET$\uparrow$} & \textbf{$\Delta$BLEU} \\
\midrule
\multicolumn{8}{c}{\textit{NMT Models}} \\
\midrule
NLLB-600M & Base & 0.77 & 14.40 & 118.13 & 0.866 & 0.484 & - \\
          & + LoRA & 5.65 & 28.74 & 88.01 & 0.900 & 0.638 & +4.89 \\ 
\midrule
NLLB-1.3B & Base & 1.25 & 16.15 & 107.03 & 0.873 & 0.525 & - \\
          & + LoRA & 5.68 & 28.94 & 88.24 & 0.900 & 0.656 & +4.43 \\ 
\midrule
M2M100-1.2B & Base & 0.07 & 9.37 & 100.34 & 0.840 & 0.428 & - \\ 
            & + QLoRA & 9.52 & 33.30 & 81.95 & \textbf{0.911} & 0.691 & \textbf{+9.45} \\ 
            & + Full FT & 8.16 & 31.12 & 83.11 & 0.907 & 0.664 & +8.09 \\ 
\midrule
\multicolumn{8}{c}{\textit{LLM}} \\
\midrule
Krikri-8B-IT & Base & 6.55 & 28.98 & 87.38 & 0.900 & 0.695 & - \\ 
             & + QLoRA & 10.37 & 34.09 & 82.28 & \textbf{0.911} & \textbf{0.717} & +3.82 \\ 
             & + Full FT & \textbf{12.80} & \textbf{35.90} & \textbf{81.40} & 0.884 & 0.716 & +6.25 \\ 
\bottomrule
\end{tabular}
\caption{Results on the Stress set (250 sentence pairs with rarer dialects). BLEU scores are BLEU-13a. $\Delta$BLEU shows improvement over the base model. Best score per metric in bold.}
\label{tab:stress_results}
\end{table*}

As expected, performance drops on the Stress set compared to the main Test set for all models, indicating the challenge posed by out-of-distribution dialects.

The Krikri-8B model maintains its leading position. The fully fine-tuned variant achieves the highest BLEU (\textbf{12.80}), chrF++ (\textbf{35.90}), and lowest TER (\textbf{81.40}) on this set. Meanwhile, the Krikri-QLoRA model matches the highest BERTScore F1 (\textbf{0.911}) and achieves the highest COMET score (\textbf{0.717}). The base Krikri model again starts from a higher baseline (BLEU 6.55) compared to the NMT models on this harder set. This suggests that the pre-training of Krikri might provide better generalization or robustness to dialectal variation within Greek.

The M2M100 models continue to show strong relative gains, with the QLoRA variant achieving a +9.45 BLEU improvement over its base. The NLLB models show more modest results, though the LoRA adaptations still vastly improve upon their near-zero base scores.

\section{Conclusion}
\label{sec:conclusion}

In this paper, we addressed the critical scarcity of resources for Ancient Greek (AG) to Modern Greek (MG) machine translation, a low-resource task compounded by the significant dialectal, historical, and genre-based diversity of the Ancient Greek source texts. Our primary contribution is the introduction of the AG-MG Parallel Corpus, the largest sentence-aligned dataset for this pair, containing 132,481 high-quality pairs. This new resource is derived from diverse literary, historical, and biblical texts and is comprehensively annotated with metadata for author, dialect, genre, and era, enabling more granular analysis in future work.

We detailed the novel hybrid alignment pipeline developed to create this corpus. This method improves upon existing approaches by first fine-tuning LaBSE embeddings on a 1,000-pair manually-aligned AG-MG subset before using them with VecAlign for initial alignment. A crucial second stage involved an LLM-based refinement step using Gemini 2.5 Flash, which was prompted to act as a "\textit{cautious data editor}" to perform misalignment detection and correction, ensuring the high quality of the final dataset.

Furthermore, we presented a comprehensive benchmark comparing modern NMT models (NLLB-600M, NLLB-1.3B, M2M100-1.2B) and a specialized Greek LLM (Llama-Krikri-8B-Instruct) on this new task. Our experiments evaluated three different adaptation strategies, including full fine-tuning, LoRA, and QLoRA. The results unequivocally confirm that fine-tuning on our corpus yields significant improvements over zero-shot performance, with BLEU scores increasing by up to +10.3 points. The fully fine-tuned Llama-Krikri-8B model achieved the highest absolute performance on the main test set, reaching a BLEU score of 13.16 and the highest chrF++ score of 34.71. Meanwhile, the QLoRA-adapted M2M100-1.2B model demonstrated the most substantial relative gains, improving from a near-zero base score to 10.96 BLEU.

The Llama-Krikri-8B model's strength was particularly evident on our Stress set, which focused on rarer AG dialects (Ionic, Doric, Homeric). On this challenging set, the fully fine-tuned Krikri model achieved the highest BLEU (12.80) and chrF++ scores, while its QLoRA-adapted variant achieved the highest COMET (0.717). This indicates superior generalization and robustness to dialectal variation compared to standard NMT models. This, combined with the fact that its full fine-tuning version performed worse on some metrics, suggests QLoRA may be a more stable and effective adaptation strategy for this LLM in this low-resource scenario.

While acknowledging limitations such as domain bias toward literary/biblical texts and the need for human evaluation, this work provides a new, high-quality dataset and a clear benchmark for a challenging task. To foster further research in Greek NLP and Digital Humanities, we will make our fine-tuned models available. However, as a precautionary measure due to the complex and often uncertain copyright status of the source materials, we are unable to release the compiled corpus directly.

\section{Limitations}
\label{sec:limitations}

While our work provides a significant new resource and benchmark, several limitations should be acknowledged. 
First, the \textbf{corpus composition} reflects the available digital sources, primarily literary, philosophical, and biblical texts. This may introduce domain bias, and models trained on it might perform less optimally on other genres. 
Second, despite our LLM-based refinement, the \textbf{alignment quality} is estimated at 95\% based on a manual review of 150 examples. While this sample indicates high quality, a larger human evaluation could provide a more robust assessment of alignment errors, particularly for highly non-literal translations or complex sentence rearrangements inherent in literary texts. 
Third, our \textbf{evaluation} primarily relies on automatic metrics. While comprehensive, these may not fully capture translation adequacy and fluency, especially for a language pair with rich morphology and stylistic variation like AG-MG. Large-scale human evaluation and error analysis would permit a deeper quality assessment. Furthermore, due to the small size of the Stress set (250 pairs), differences in metrics like BLEU between models may lack statistical reliability, and confidence intervals should be considered in future evaluations.
Fourth, our \textbf{modeling experiments} explored specific architectures and PEFT methods. Other models or fine-tuning techniques might yield different results. 

Notably, this work focused exclusively on \textbf{sentence-level translation}. Our original data sources were often excerpt-based, and we retained this paragraph-level alignment separately. Future work should explore leveraging this longer-context data, potentially by fine-tuning LLMs like Krikri-8B for document-level AG-MG translation, which might better handle discourse phenomena and context-dependent translations, especially for literary texts \citep{karpinska2023large, wang2023document}. Analyzing model performance across different dialects and genres using the corpus metadata is another area for future research. Finally, our experiments focused solely on the AG$\rightarrow$MG translation direction. While our parallel corpus could potentially be reversed to train MG$\rightarrow$AG models, generating morphologically rich and syntactically complex Ancient Greek poses significantly greater challenges, representing a distinct direction for future investigation.

\section{Ethical Considerations}
\label{sec:ethics}

The data used in this work was compiled from publicly accessible online resources, primarily intended for educational and research purposes. We believe our use aligns with the intended purpose of these digital libraries. The created dataset, AG-MG Parallel Corpus, consists of historical texts and their modern translations. While the source texts themselves may reflect historical biases, our processing did not introduce new biases beyond potential selection effects from the source availability. 

The models fine-tuned are based on open-source pre-trained models (NLLB, M2M100, Llama-Krikri). Our intended use for the fine-tuned models and the corpus is for research, education, and cultural heritage preservation within the NLP and Digital Humanities communities. Potential misuse seems minimal given the specific language pair and domain, but users should be aware that, although MT quality is high according to metrics, it is not perfect and should not be relied upon for critical applications without human oversight.

\section{Acknowledgments}
\label{sec:acknowledgments}

This work was supported in part by a thesis scholarship granted to the first author by the Institute for Language and Speech Processing (ILSP), Athena Research Center. This work was also supported in part by the PHAROS project (Grant Agreement No. 101234269). We acknowledge the EuroHPC Joint Undertaking for awarding this project access to the LEONARDO supercomputer, hosted by CINECA (Italy) and the LEONARDO consortium through the EuroHPC Development Access call.

\section{Bibliographical References}\label{sec:reference}

\bibliographystyle{lrec2026-natbib}
\bibliography{lrec2026}

\section{Language Resource References}
\label{lr:ref}
\bibliographystylelanguageresource{lrec2026-natbib}
\bibliographylanguageresource{languageresource}

\appendix
\titleformat{\section}{\normalfont\large\bfseries\center}{Appendix \thesection.}{0.5em}{}

\section{Corpus Creation Pipeline}
\label{app:pipeline}

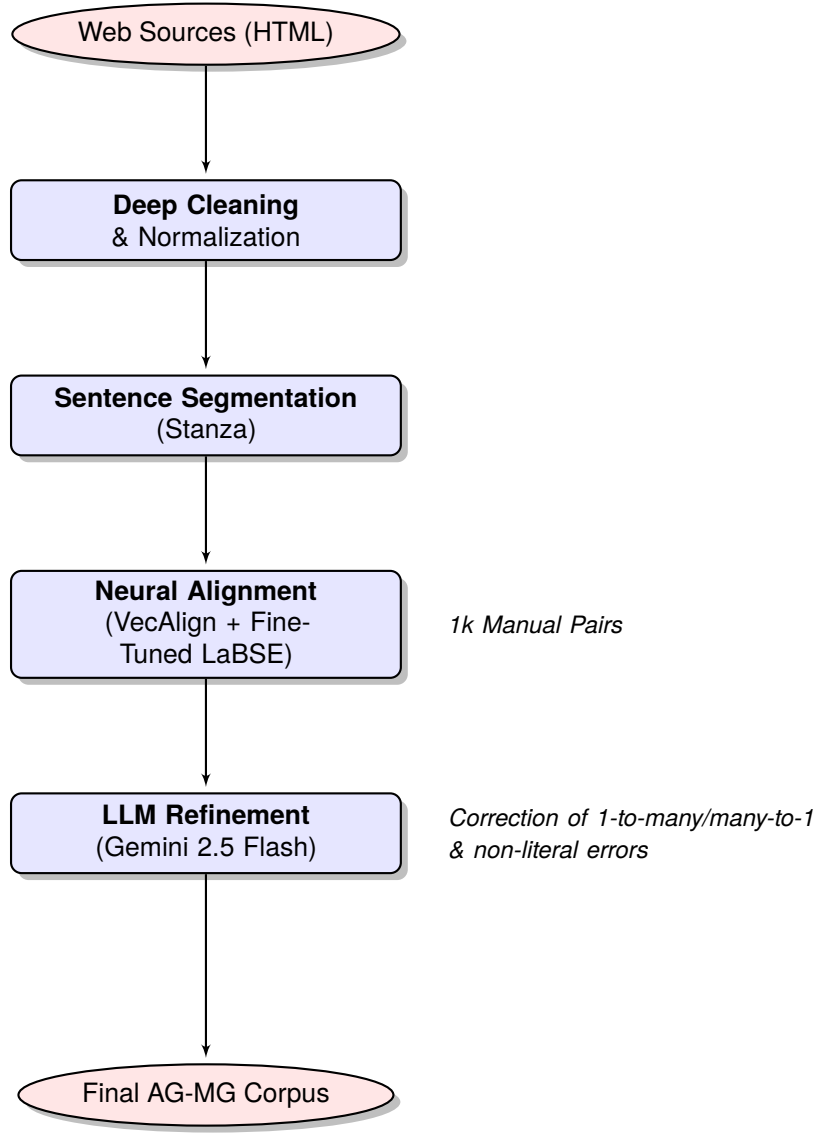
\begin{figure*}[!ht]
\centering
\begin{tikzpicture}[node distance=1.5cm, auto,
    block/.style={rectangle, draw=black, thick, fill=blue!10, text width=14em, text centered, rounded corners, minimum height=3em, drop shadow},
    cloud/.style={draw=black, thick, ellipse, fill=red!10, node distance=2.5cm, minimum height=2em, drop shadow},
    line/.style={draw, thick, -latex', shorten >=2pt}]

    \node [cloud] (input) {Web Sources (HTML)};
    \node [block, below=of input] (clean) {\textbf{Deep Cleaning} \\ \& Normalization};
    \node [block, below=of clean] (segment) {\textbf{Sentence Segmentation} \\ (Stanza)};
    \node [block, below=of segment] (align) {\textbf{Neural Alignment} \\ (VecAlign + Fine-Tuned LaBSE)};
    \node [block, below=of align] (refine) {\textbf{LLM Refinement} \\ (Gemini 2.5 Flash)};
    \node [cloud, below=of refine] (output) {Final AG-MG Corpus};

    \path [line] (input) -- (clean);
    \path [line] (clean) -- (segment);
    \path [line] (segment) -- (align);
    \path [line] (align) -- (refine);
    \path [line] (refine) -- (output);
    
    \node [text width=5cm, right=0.5cm of align] {\small \textit{1k Manual Pairs}};
    \node [text width=5cm, right=0.5cm of refine] {\small \textit{Correction of 1-to-many/many-to-1 \& non-literal errors}};

\end{tikzpicture}
\caption{The hybrid corpus creation pipeline, combining neural embedding-based alignment with LLM-based refinement.}
\label{fig:pipeline}
\end{figure*}

Figure \ref{fig:pipeline} illustrates the multi-stage hybrid alignment pipeline used to create the AG-MG Parallel Corpus, combining neural embeddings with LLM-based refinement.

\section{Detailed Corpus Statistics}
\label{app:statistics}

This section provides a granular breakdown of the AG-MG Parallel Corpus. Tables \ref{tab:split_train_dialects_AG_MG} through \ref{tab:split_stress_dialects_AG_MG} detail the dialect distributions across the specific data splits. 

\begin{figure*}[htbp]
    \centering
    \includegraphics[width=0.85\textwidth]{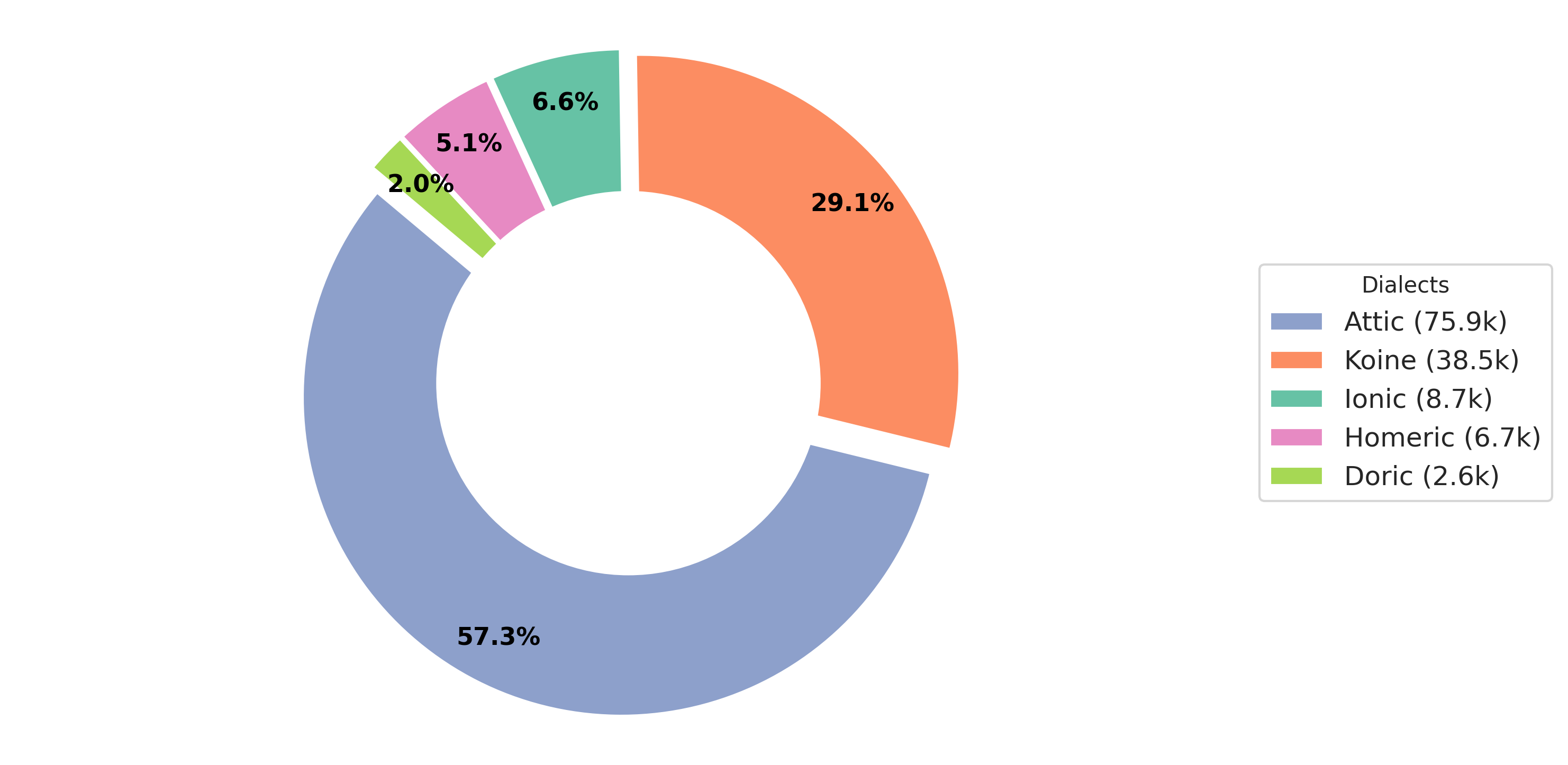}
    \caption{Distribution of Ancient Greek dialects in the sentence-level AG$\rightarrow$MG corpus. The dataset is predominantly Attic (57.3\%) and Koine (29.1\%), reflecting the availability of digital resources.}
    \label{fig:dialects_pie}
\end{figure*}

\begin{figure*}[htbp]
    \centering
    \includegraphics[width=0.85\textwidth]{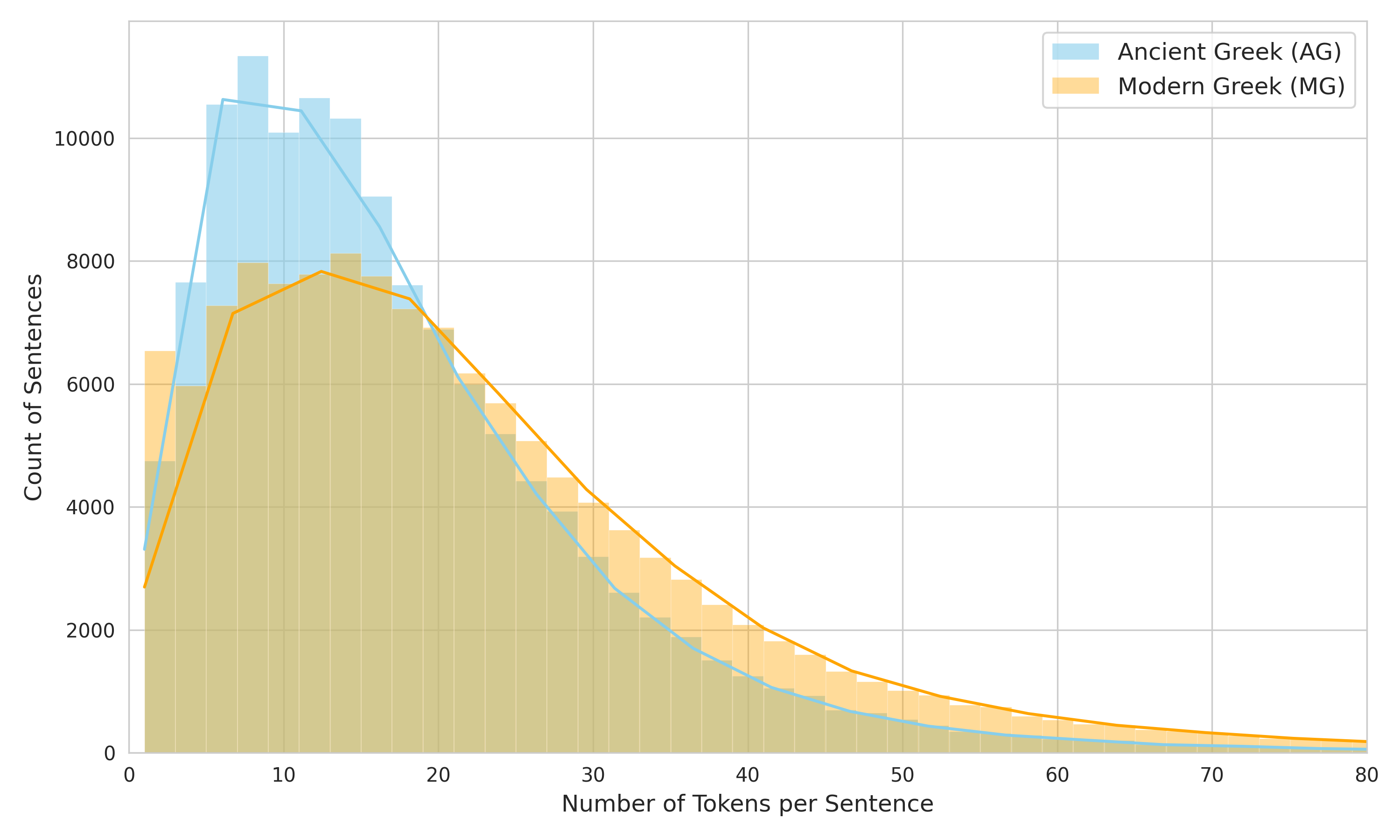}
    \caption{Sentence-Level token count distribution (132.4k pairs). Most sentences are short (10-30 tokens), with Modern Greek translations showing a slightly wider distribution due to their analytic nature.}
    \label{fig:token_distributions_AG_MG}
\end{figure*}

Figure \ref{fig:dialects_pie} visualizes the overall dominance of the Attic and Koine dialects. Finally, as shown in Table \ref{tab:stats}, the corpus exhibits a significant token count disparity between the source and target languages. The Modern Greek side contains approximately 32\% more tokens than the Ancient Greek side (3.06M vs. 2.31M). This expansion is linguistically expected and reflects the typological shift from Ancient Greek, a highly synthetic language rich in inflection, to Modern Greek, which is more analytic. Ancient Greek often encodes grammatical information (such as subject, tense, and mood) into single word endings, whereas Modern Greek frequently employs periphrastic constructions, auxiliary verbs, and particles to convey the same meaning. This is also visually evident in the token distribution histogram (Figure \ref{fig:token_distributions_AG_MG}), where the Modern Greek distribution is consistently shifted towards higher token counts compared to the Ancient Greek counterpart.

\begin{table}[!ht]
\centering
\small
\begin{tabular}{l rr}
\toprule
\textbf{Dialect} & \textbf{Rows} & \textbf{Pairs} \\
\midrule
Attic & 3,089 & 73,287 \\
Ionic & 232 & 8,615 \\
Doric & 244 & 2,554 \\
Homeric / Epic & 272 & 6,649 \\
Hellenistic Koine & 1,491 & 37,126 \\
\midrule
\textbf{Total} & \textbf{5,328} & \textbf{128,231} \\
\bottomrule
\end{tabular}
\caption{Sentence-Level Dialect Distribution: Train Set.}
\label{tab:split_train_dialects_AG_MG}
\end{table}

\begin{table}[!ht]
\centering
\small
\begin{tabular}{l rr}
\toprule
\textbf{Dialect} & \textbf{Rows} & \textbf{Pairs} \\
\midrule
Attic & 134 & 1,300 \\
Hellenistic Koine & 58 & 700 \\
\midrule
\textbf{Total} & \textbf{192} & \textbf{2,000} \\
\bottomrule
\end{tabular}
\caption{Sentence-Level Dialect Distribution: Dev Set.}
\label{tab:split_dev_dialects_AG_MG}
\end{table}

\begin{table}[!ht]
\centering
\small
\begin{tabular}{l rr}
\toprule
\textbf{Dialect} & \textbf{Rows} & \textbf{Pairs} \\
\midrule
Attic & 48 & 1,300 \\
Hellenistic Koine & 28 & 700 \\
\midrule
\textbf{Total} & \textbf{76} & \textbf{2,000} \\
\bottomrule
\end{tabular}
\caption{Sentence-Level Dialect Distribution: Test Set.}
\label{tab:split_test_dialects_AG_MG}
\end{table}

\begin{table}[H]
\centering
\small
\begin{tabular}{l rr}
\toprule
\textbf{Dialect} & \textbf{Rows} & \textbf{Pairs} \\
\midrule
Ionic & 3 & 110 \\
Doric & 9 & 60 \\
Homeric / Epic & 5 & 80 \\
\midrule
\textbf{Total} & \textbf{17} & \textbf{250} \\
\bottomrule
\end{tabular}
\caption{Sentence-Level Dialect Distribution: Stress Set.}
\label{tab:split_stress_dialects_AG_MG}
\end{table}

\section{Qualitative Translation Examples}
\label{app:qualitative}

Table \ref{tab:qual_example_test} provides a representative qualitative comparison of model outputs on a sentence from the Test set. The base NMT models often fail or hallucinate unrelated text, while the fine-tuned versions align correctly with the context and vocabulary.

\begin{table*}[!ht]
\centering
\renewcommand{\arraystretch}{1.4}
\small
\begin{tabular}{@{} p{0.18\textwidth} p{0.78\textwidth} @{}}
\toprule
\textbf{Model} & \textbf{Text} \\
\midrule
\textbf{Source (AG)} & {\greekfont Ἐκ δὲ τούτου Παρμενίωνα μὲν πέμπει ἐπὶ τὰς ἄλλας πύλας, αἳ δὴ ὁρίζουσι τὴν Κιλίκων τε καὶ Ἀσσυρίων χώραν, προκαταλαβεῖν καὶ φυλάσσειν τὴν πάροδον, δοὺς αὐτῷ τῶν τε ξυμμάχων τοὺς πεζοὺς καὶ τοὺς Ἕλληνας τοὺς μισθοφόρους καὶ τοὺς Θρᾷκας, ὧν Σιτάλκης ἡγεῖτο, καὶ τοὺς ἱππέας δὲ τοὺς Θεσσαλούς.} \\
\midrule
\textbf{NLLB-600M (Base)} & {\greekfont Μέχρι να περάσει ο Παρμηνίος από τις άλλες πύλες, που βρίσκονται στην περιοχή της Κιλικνίας και της Ασσυρίας, να προχωρήσει και να φυλάξει το πέρασμα, να κατασκευάσει τα τετράγωνα των πεζοδρόμων, και τα ελληνικά μισθοφόρα και τα θραύσματα, και το Σιτάλκιο και το Σπέπας των Θεσσαλονίκων.} \\
\textbf{NLLB-600M (LoRA)} & {\greekfont Ύστερα έστειλε τον Παρμενίωνα στις άλλες πύλες, που καθορίζουν τη χώρα των Κιλίκων και τωνἎσσυρίων, για να προλάβει και να φυλάξει το πέρασμα, δίνοντάς του και τους πεζούς των συμμάχων του και τωνἘλλήνων τους μισθοφόρους και τους Θράκες, που τους διοικούσε ο Σιτάλκης, και τα άλογα των Θεσσαλών.} \\
\midrule
\textbf{NLLB-1.3B (Base)} & {\greekfont Ο Παρμαινίος με έστειλε στην Ελληνική πύλη, στις περιοχές της Κιλικίας και της Συρίας, για να προχωρήσω και να προστατέψω το ταξίδι, με τους πεζούς ναυτικούς, τους Ελληνες ναυτικούς, τον Τραύκο, τον Στάλκη, τον Γεώργιο και τον Ιάππο στη Θεσσαλονίκη.} \\
\textbf{NLLB-1.3B (LoRA)} & {\greekfont Ύστερα απ ̓ αυτό έστειλε τον Παρμενίωνα στις άλλες πύλες, που χωρίζουν τη χώρα των Κιλικόνων και τωνἊσσυρίων, για να προλάβει και να φυλάξει το πέρασμα, και του παρέδωσε το πεζικό των συμμάχων, τους Έλληνες μισθοφόρους και τους Θράκες, που τους οδηγούσεὀ Σίλκης, και τους Θεσσαλούςἱππείς.} \\
\midrule
\textbf{M2M-1.2B (Base)} & {\greekfont \textit{[Hallucination]} ταν ὁ ἄνθρωπος εἶναι ἄλλος, ὅταν ἔχει ἐντολή νά ὑπάρχει ἀπὸ τήν μαρτία του, ἐκείνος ἐπικρατεί ὅτι ἐξαρτάται ἀπό τόν αυτό του.} \\
\textbf{M2M-1.2B (QLoRA)} & {\greekfont Ύστερ ̓ απ ̓ αυτό έστειλε τον Παρμενίωνα στις άλλες πύλες, που ορίζουν τη χώρα των Κιλίκων και των Ασσυρίων, για να προλάβουν και να φρουρήσουν το πέρασμα, δίνοντάς του από τους συμμάχους τους πεζούς, τους Έλληνες τους μισθοφόρους και τους Θράκες, με αρχηγό τον Σιτάλκη, και το ιππικό τους τους Θεσσαλούς.} \\
\textbf{M2M-1.2B (FFT)} & {\greekfont Ύστερ ̓ απ ̓ αυτό έστειλε τον Παρμενίωνα στις άλλες πύλες, που ορίζουν τη χώρα των Κιλίκων και των Ασσυρίων, για να προλάβει και να φυλάξει το πέρασμα, δίνοντάς του τους πεζούς των συμμάχων, τους Έλληνες, τους μισθοφόρους, τους Θράκες, που τους οδηγούσε ο Σιτάλκης, και τους ιππείς τους Θεσσαλούς.} \\
\midrule
\textbf{Krikri-8B-IT (Base)} & {\greekfont Μετά το γεγονός αυτό, ο Παρμενίων στάλθηκε στις υπόλοιπες πύλες που οριοθετούν την περιοχή της Κιλικίας και της Ασσυρίας, με εντολή να προλάβει και να φυλάξει τις διαβάσεις, δίνοντάς του τους πεζούς συμμάχους, τους Έλληνες μισθοφόρους και τους Θράκες (με αρχηγό τον Σιτάλκη) καθώς και τους Θεσσαλούς ιππείς.} \\
\textbf{Krikri-8B-IT (QLoRA)} & {\greekfont Ύστερα απ᾽ αυτό ο Αλέξανδρος έστειλε τον Παρμενίωνα να καταλάβει και να φρουρήσει τις άλλες πύλες που χωρίζουν τη χώρα των Κιλίκων και των Ασσυρίων, δίνοντάς του τους πεζούς συμμάχους και τους Έλληνες μισθοφόρους και τους Θράκες, που τους διοικούσε ο Σιτάλκης, καθώς και το θεσσαλικό ιππικό.} \\
\textbf{Krikri-8B-IT (FFT)} & {\greekfont Μετά από αυτό έστειλε τον Παρμενίωνα στις άλλες πύλες, με τις οποίες ορίζεται η χώρα των Κιλίκων και των Ασσυρίων, για να καταλάβει και να φυλάξει τα περάσματα, δίνοντάς του τους πεζούς των συμμάχων, τους Έλληνες και τους Θράκες μισθοφόρους, τους οποίους διοικούσε ο Σιτάλκης, καθώς και το θεσσαλικό ιππικό} \\
\bottomrule
\end{tabular}
\caption{Qualitative comparison on a Test Set sentence (\textit{The Anabasis of Alexander}, Arrian, Attic dialect). Some base models hallucinate or fail, while fine-tuned models produce fluent Greek.}
\label{tab:qual_example_test}
\end{table*}

\end{document}